\documentclass[twoside,11pt]{article}

\usepackage{subcaption}
\usepackage{amsmath}

\usepackage{jmlr2e}

\ShortHeadings{The Human Visual System and Adversarial AI}{Yaoshiang Ho and Samuel Wookey}
\firstpageno{1}

\begin{document}

\title{The Human Visual System and Adversarial AI}

\author{\name Yaoshiang Ho \email yaoshiang@thinky.ai \\
       \addr Thinky.AI Research \\
       Los Angeles, CA USA
       \AND
       \name Samuel G.\ Wookey \email sam@thinky.ai \\
       \addr Thinky.AI Research \\
       Los Angeles, CA USA
    }

\editor{}

\maketitle

\begin{abstract}
This paper applies theories about the Human Visual System to make
Adversarial AI more effective. To date, Adversarial AI has modeled perceptual distances between clean and adversarial
examples of images using $L_p$ norms. These norms have the benefit of 
simple mathematical description and reasonable effectiveness in approximating perceptual distance. 
However, in prior decades, other areas of
image processing have moved beyond simpler models like Mean Squared Error (MSE) towards
more complex models that better approximate the Human Visual System (HVS). We demonstrate a 
proof of concept of incorporating HVS models into Adversarial AI. 
\end{abstract}

\begin{keywords}
  Adversarial AI, Human Visual System, Discrete Cosine Transformation, Luma, Chroma
\end{keywords}

\section{Introduction}

Adversarial AI is a set of techniques to find a small change to an input that nevertheless changes its classification by a classifier. The initial research that launched the field focused on images as the input and deep convolutional neural networks (DCNN) as the classifier \citep{sze2013}. 

A key to Adversarial AI is the minimization of the perceptual distance of the changes. Obviously, an unbounded change would trivially completely replace an image of say a truck with a horse. The central challenge of Adversarial AI is to change the pixels of an image of say a truck while minimizing perceptual distance, so that a human would still easily classify it as an image of a truck but a DCNN would be confused into classifying it as a horse. 

The minimization of this change is often measured by $L_p$ norms. An $L_0$ norm counts the number of pixels changed. An $L_1$ norm sums up the magnitude of the change over all pixels. An $L_2$ norm is the square root of the sum of the squares of the changes. And the $L_\infty$ norm is the magnitude of the most changed pixel. \citet{car2017} created multiple figures that demonstrate the differences between adversarial images generated using each of these norms. 

In this paper, we explore alternative perceptual distance metrics based on understandings of the human visual system (HVS). The HVS encompasses the biochemical, neurological, and psychological models of human vision. HVS is often applied to find more effective methods of image and video compression. In image compression, there has already been a shift from simpler models like Mean Squared Error (MSE) towards more complex models that better approximate the HVS \citep{wang2002image}. 

We draw on two basic theories about the HVS and find that they lead to more effective generation of adversarial data in some cases. 

\begin{figure}[t]
    \centering
    \begin{subfigure}{0.45\textwidth}
        \includegraphics[width=\textwidth]{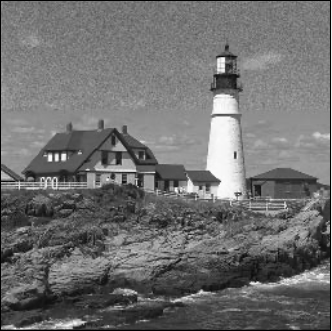}
        \caption{Perturbations applied to the low frequency area of the sky.}
        \label{fig:masksky}
    \end{subfigure}
    ~ 
    \begin{subfigure}{0.45\textwidth}
        \includegraphics[width=\textwidth]{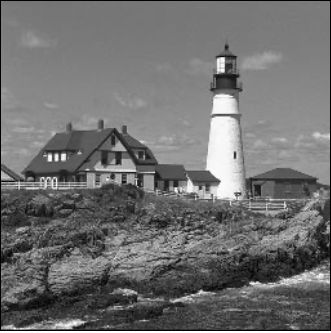}
        \caption{Perturbations applied to the high frequency area of the rocks.}
        \label{fig:maskrock}
    \end{subfigure}
    ~ 
    \caption{Example of masking in high frequencies. Identical amounts of noise have been added to both images. The perturbations in the low frequency area of the sky (left) is more noticeable than the perturbations in the high frequency area of the rocks (right) \citep[Fig. 2]{nadenau2000human}. }
    \label{fig:pert}
\end{figure}

The first concept is that the HVS is more sensitive to lower frequency information (Figure \ref{fig:pert}). This is the basis for the discrete cosine transform (DCT) methodology of lossy image compression \citep{JPEG}. 

\begin{figure}
    \centering
    \begin{subfigure}[t]{0.23\textwidth}
        \includegraphics[width=\textwidth]{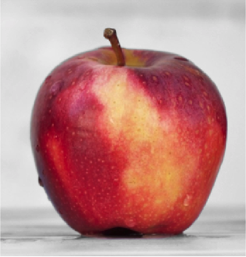}
        \caption{A clean color image of an apple.}
        \label{fig:appleclean}
    \end{subfigure}
    ~ 
    \begin{subfigure}[t]{0.23\textwidth}
        \includegraphics[width=\textwidth]{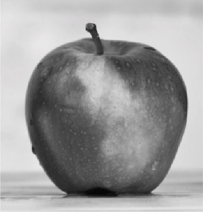}
        \caption{Black and white images retain the luma but eliminate chroma information.}
        \label{fig:applebw}
    \end{subfigure}
    ~ 
    \begin{subfigure}[t]{0.23\textwidth}
        \includegraphics[width=\textwidth]{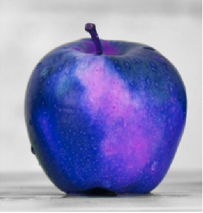}
        \caption{An image with heavy distortion of chroma. }
        \label{fig:appleblue}
    \end{subfigure}
    ~
    \begin{subfigure}[t]{0.23\textwidth}
        \includegraphics[width=\textwidth]{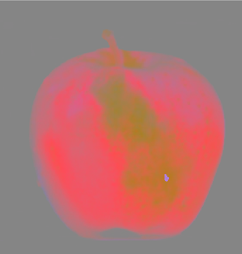}
        \caption{An image with chroma unchanged but luma brought to a constant level.}
        \label{fig:appleluma}
    \end{subfigure}
    \caption{Example of luma and chroma importance. Retaining luma while eliminating or distorting chroma information results in an image that is still easily identifiable as an apple. Adjusting luma to a constant level while leaving chroma unchanged results in an image that is more difficult to classify \citep{zeileis2019colorspace}. }
    \label{fig:apple}
\end{figure}

The second concept is that the HVS is more sensitive to changes in luma (brightness) than chroma (hue) (Figure \ref{fig:apple}). This was discovered by \citet{bedford1950mixed} during pioneering work on color television, and downsampled chroma channels continue to exist in standards today like MPEG and Apple ProRes 422 \citep{prores}.

Our contribution is as follows. We define our problem as reducing the perceptual distance of adversarial attacks on images. Our approach is to apply HVS concepts better target the location and perturbation of adversarial attacks. In practice, this could help an adversary better hide their adversarial attacks from manual detection. 

We combine two approaches into what we call the HVS2 attack. Our first approach is to only perturb pixels in high frequency zones. We use a simple model of high frequency. Our second approach is to perturb pixels to retain an approximately constant chroma. We find that these pixels are in the same "color palette" as the rest of the image, making them less detectable. 

The HVS2 attack works well in some cases and poorly in others.

\section{Background and Related Work}

Since its creation, the field of Adversarial AI has acknowledged the role of the HVS. In the original work of \citet{sze2013} that launched the field of Adversarial AI, the authors describe adversarial images as "visually hard to distinguish". 

More recently, \citet{car2017} called for work into additional models for perceptual distance: "$L_p$ norms are reasonable approximations of human perceptual distance [...] No distance metric is a perfect measure of human perceptual similarity, and we pass no judgement on exactly which distance metric is optimal. We believe constructing and evaluating a good distance metric is an important research question we leave to future work."

In the first adversarial attack, several design choices were made. The DCNN to be attacked was analyzed by an adversary in a "whitebox" setting, meaning that the attack had access to internal values of the model not normally accessible to end users. The distance metric used was $L_2$ norm. The input data type were images, and the specific optimization to discover the adversarial example was L-BFGS. The perturbations affected specific pixels. 

Subsequent attacks have expanded the design space.

Additional optimization methods were applied, including fast gradient sign method (FGSM) \citep{goodfellow2014explaining}, basic iterative method (BIM) \citep{kurakin2016adversarial}, and projected gradient descent (PGD) \citep{madry2017towards}. 

The multiple options for norms were demonstrated by \citet{car2017}. \citet{warde201611} argue that the L$\infty$ norm is a preferable choice for distance metric. 

Additional secrecy models were introduced and attacked. If the internal values of the model are hidden, then a model can be attacked by an enhancement of finite differences called simultaneous perturbation stochastic approximation (SPSA) \citep{uesato2018adversarial} and boundary attack \citep{brendel2017decision}. Hiding even the output of the softmax behind a top-1 hard-labeling function can be attacked \citep{cheng2018query}. Other non-differentiable layers can be approximated by backward pass differentiable approximation (BPDA) \citep{athalye2018obfuscated}.

Input types were expanded to include audio, text, and structured data \citep{cheng2018query}.

The classifier designs expanded beyond DCNN to include RNN, SVM, and gradient boosted decision trees \citep{papernot2016transferability}. 

A series of defenses have been proposed, including defensive distillation \citep{papernot2016effectiveness}. Nearly all have been defeated except the original approach proposed: adversarial retraining \citep{uesato2018adversarial} and an optimization, logit pairing \citep{kannan2018adversarial}. 

Finally, the attacks have moved beyond pixel-perfect attacks into the physical world \citep{kurakin2016adversarial, eykholt2017robust, tencent}.

The term adversarial is also used in the field of generative adversarial nets (GANs) \citep{goodfellow2014generative}. The term adversarial is used identically in both GANs and Adversarial AI. In both fields, a generator attempts to create adversarial examples that fool a discriminator. In a GAN, the generator and discriminator are two separate deep neural networks (DNN). By comparison, in Adversarial AI the discriminator is an existing classifier and the generator is a hijacked version of the classifier itself, rather than a separate DNN. If an analogy for GANs is a counterfeiter and a cop competing with each other, the analogy for Adversarial AI is that the cop is brainwashed into using its police skills to create counterfeits. 

\section{The HVS2 Attack}

We chose the generic DCNN architecture specified in the Keras documentation for it small computational complexity. We ran our attack on 100 images from CIFAR10 \citep{cifar10}. We used ourselves as the human subjects to qualitatively measure perceptual distance.

\begin{table}[t]
\begin{center}
\begin{tabular}{ c c } 
 \hline
 Layer Type & Hyperparameters \\
 \hline
 Convolution + Relu & 32x3x3 \\
 Convolution + Relu & 32x3x3 \\
 Maxpool & 2x2 \\
 Dropout & 0.25 \\
 Convolution + Relu & 64x3x3 \\
 Convolution + Relu & 64x3x3 \\
 Maxpool & 2x2 \\
 Dropout & 0.25 \\
 Flatten & \\
 Dense & 512 \\
 Dropout & 0.5 \\
 Dense + Softmax & 10 \\
 \hline
\end{tabular}
\end{center}
\caption{Our DCNN architecture. }
\label{table:dcnn}
\end{table}

We based our attacks on the FGSM method for its low computational requirements. For ease of implementation, we implemented FGSM ourselves rather than modify the reference implementation in cleverhans \citep{papernot2016technical}.

Our HSV2 attack combines hiding perturbations in high frequency areas and constant chroma. 

For hiding perturbations in high frequency areas, we built our own simple measure of high frequency. For each pixel's color channel, we calculated two means: the mean of the above and below pixel's color channels, and the mean of the left and right color channels. We ignored pixels on the edges and corners. 

With the vertical mean and horizontal mean in hand, we calculated the absolute value of the pixel channel's deviation from each of these means. Then we took the min of those two deviations. This is our approximation of frequency for each color channel per pixel. 

For each pixel, we took the max of the frequencies of the three color channels. This is our approximation of frequency for a pixel. We reasoned that even if the channel frequencies from say red and green were low, a high frequency for the blue channel would still cause the HVS to perceive high frequency for the pixel. 

With an estimate of frequency for each individual pixel, we only allowed FGSM to adjust pixels that had a higher than a specific threshold. Any pixel with a lower frequency was not perturbed. We tried several thresholds and found 0.01 to generate reasonable results for some images. 

As described below, our initial hypothesis of constant luma failed to product effective results but led us to the approach of constant chroma. We approximated constant chroma by only allowing FGSM to operate on pixels where the sign of the gradients on all three color channels were either positive or negative. We reason that constant chroma creates perturbed pixels within the same color palette as the rest of the image, reducing perceptual distance. 

\begin{figure}[p]
    \centering
    \begin{subfigure}[h]{0.3\textwidth}
        \begin{center}
        \includegraphics[width=\columnwidth]{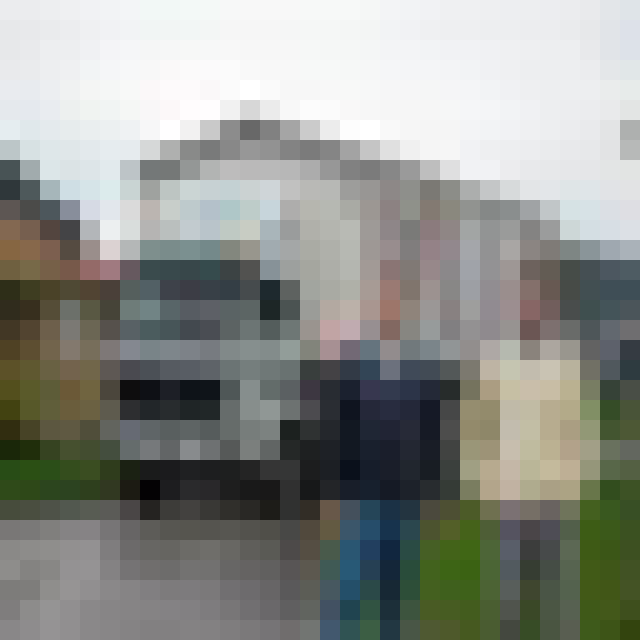}
        \end{center}
    \end{subfigure}
    ~ 
    \begin{subfigure}[h]{0.3\textwidth}
        \begin{center}
        \includegraphics[width=\columnwidth]{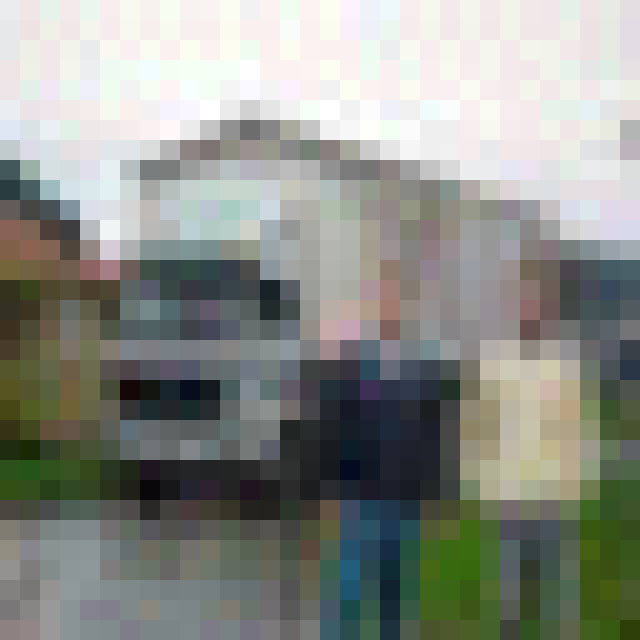}
        \end{center}
    \end{subfigure}
    ~ 
    \begin{subfigure}[h]{0.3\textwidth}
        \begin{center}
        \includegraphics[width=\columnwidth]{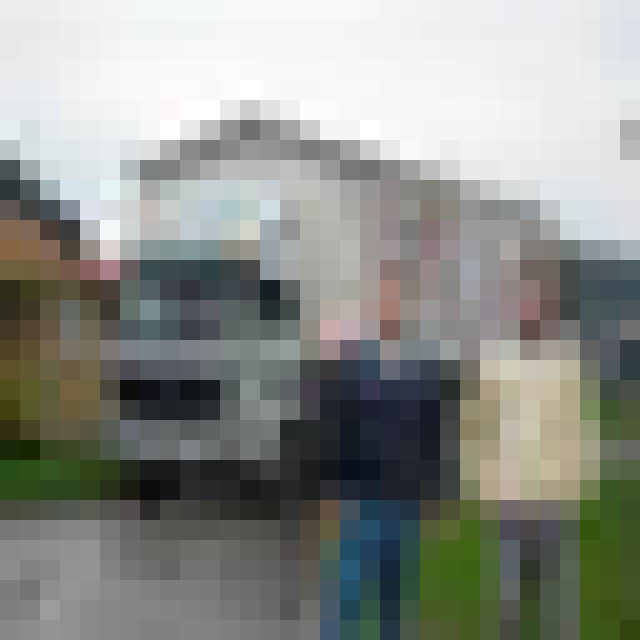}
        \end{center}
    \end{subfigure}
    \begin{subfigure}[h]{0.3\textwidth}
        \begin{center}
        \includegraphics[width=\columnwidth]{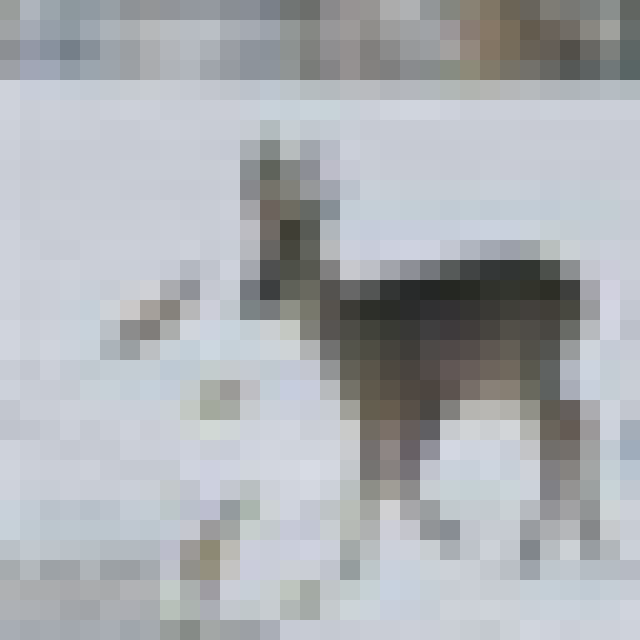}
        \end{center}
    \end{subfigure}
    ~ 
    \begin{subfigure}[h]{0.3\textwidth}
        \begin{center}
        \includegraphics[width=\columnwidth]{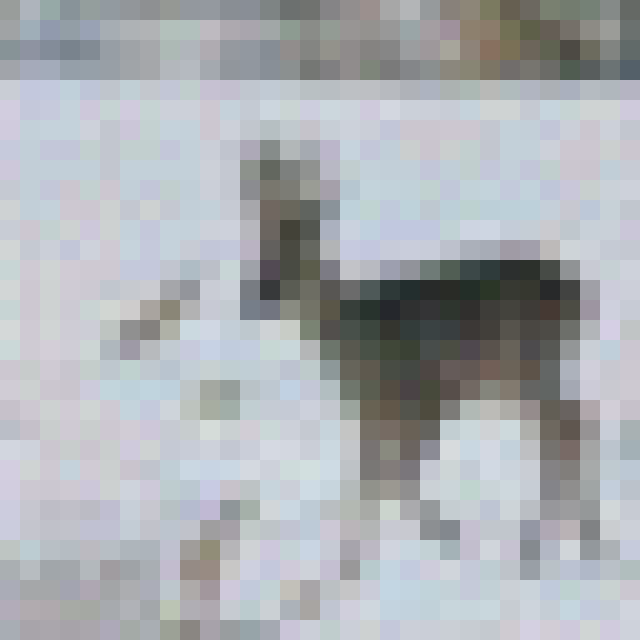}
        \end{center}
    \end{subfigure}
    ~ 
    \begin{subfigure}[h]{0.3\textwidth}
        \begin{center}
        \includegraphics[width=\columnwidth]{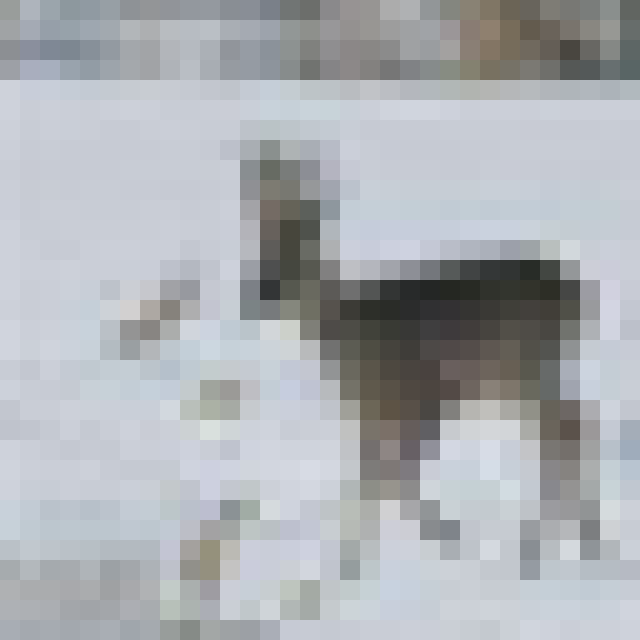}
        \end{center}
    \end{subfigure}
    \begin{subfigure}[h]{0.3\textwidth}
        \begin{center}
        \includegraphics[width=\columnwidth]{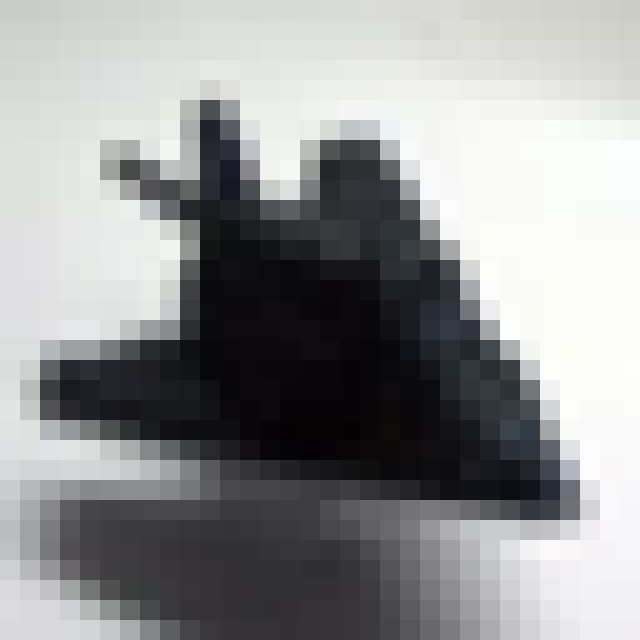}
        \end{center}
        \caption{Original images.}
    \end{subfigure}
    ~ 
    \begin{subfigure}[h]{0.3\textwidth}
        \begin{center}
        \includegraphics[width=\columnwidth]{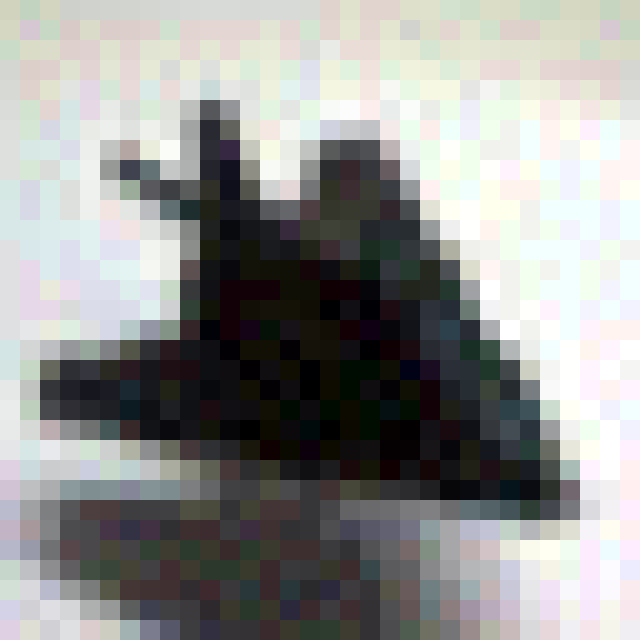}
        \end{center}
        \caption{FGSM attack.}
    \end{subfigure}
    ~ 
    \begin{subfigure}[h]{0.3\textwidth}
        \begin{center}
        \includegraphics[width=\columnwidth]{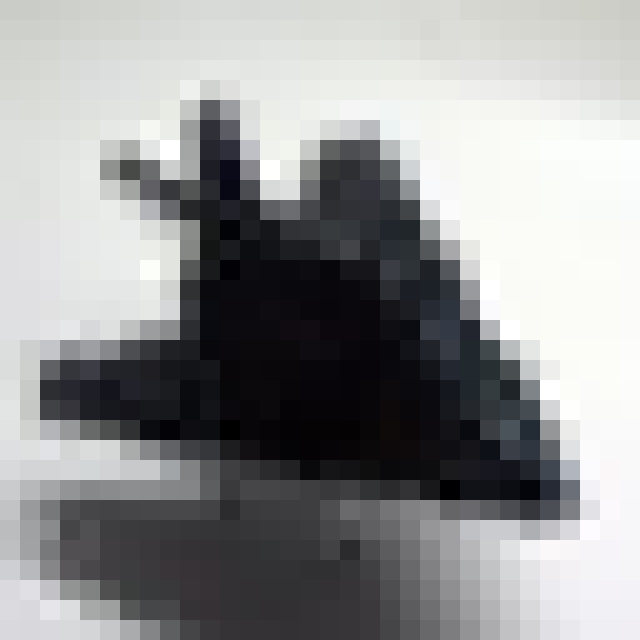}
        \end{center}
        \caption{HVS2 attack. }
        \label{bad8}
    \end{subfigure}
    \caption{The good results. The clean images in column (a) show "smooth" low frequency regions. The FGSM attacked images in column (b) show "rainbow snow" in those regions. The HVS2 attacked images in column (c) reduce chroma changes and hide adversarial pixels in high frequency areas, leading to lower perceptual distance. Original image size 32x32. }
    \label{fig:good}
\end{figure}

\begin{figure}[p]
    \centering
    \begin{subfigure}[h]{0.3\textwidth}
        \begin{center}
        \includegraphics[width=\columnwidth]{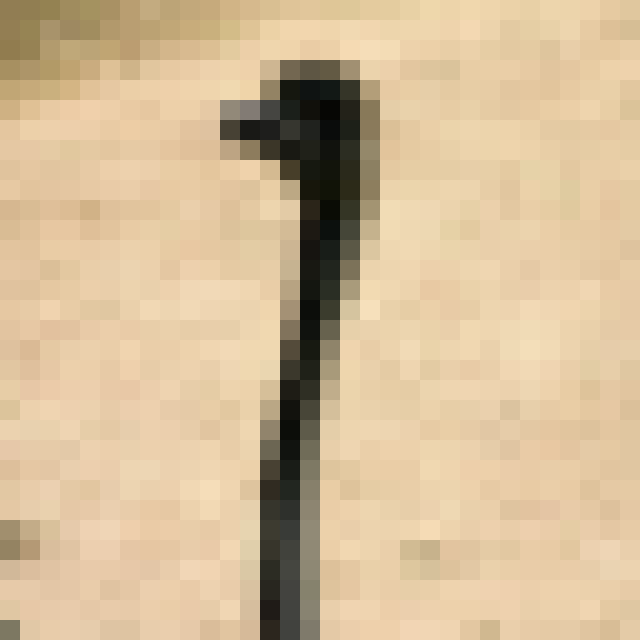}
        \end{center}
    \end{subfigure}
    ~ 
    \begin{subfigure}[h]{0.3\textwidth}
        \begin{center}
        \includegraphics[width=\columnwidth]{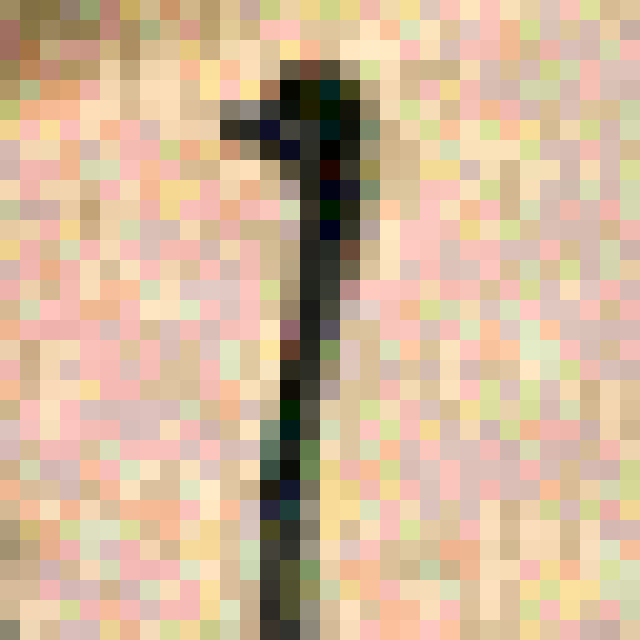}
        \end{center}
    \end{subfigure}
    ~ 
    \begin{subfigure}[h]{0.3\textwidth}
        \begin{center}
        \includegraphics[width=\columnwidth]{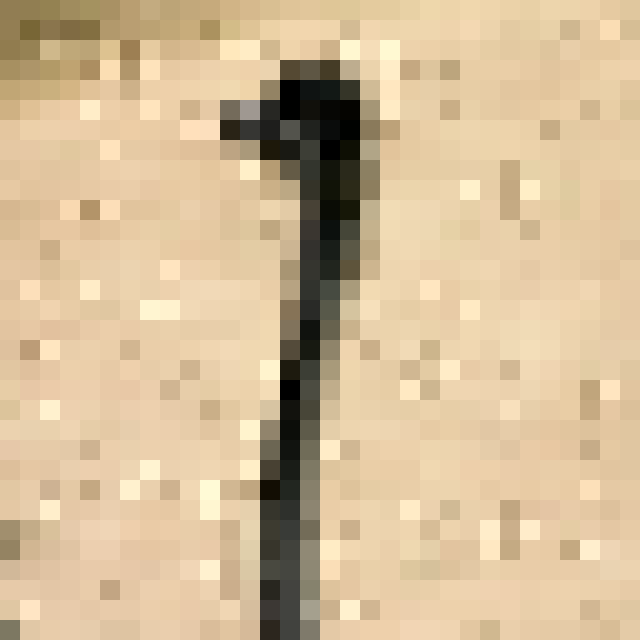}
        \end{center}
    \end{subfigure}
    \begin{subfigure}[h]{0.3\textwidth}
        \begin{center}
        \includegraphics[width=\columnwidth]{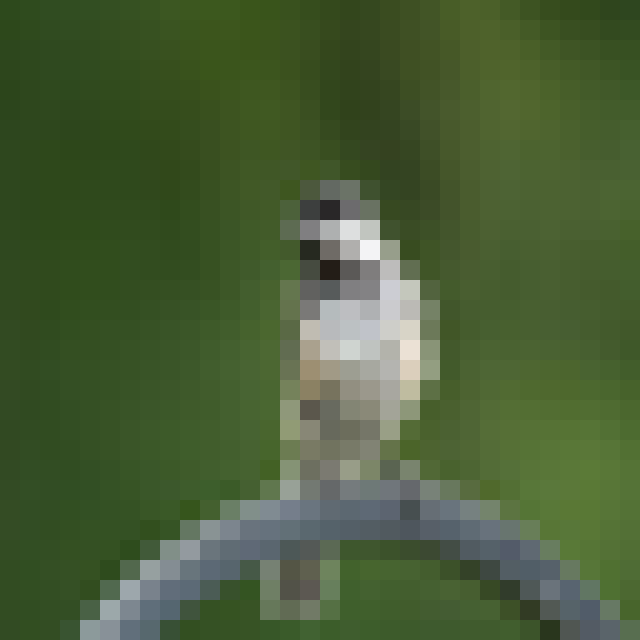}
        \end{center}
    \end{subfigure}
    ~ 
    \begin{subfigure}[h]{0.3\textwidth}
        \begin{center}
        \includegraphics[width=\columnwidth]{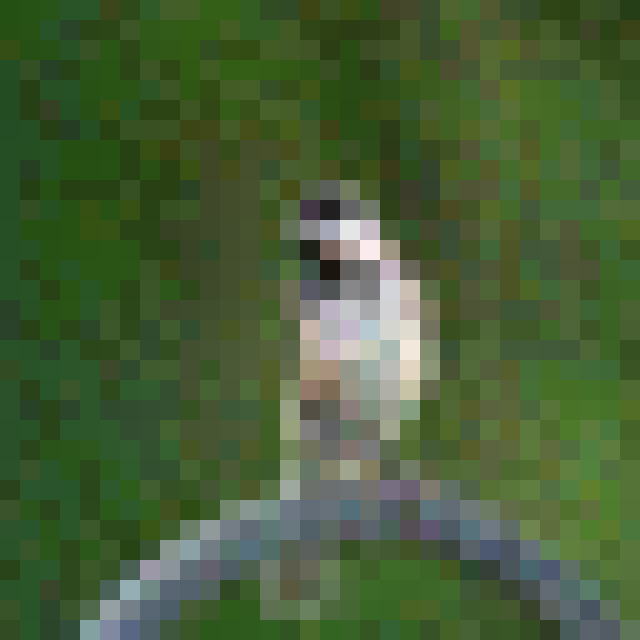}
        \end{center}
    \end{subfigure}
    ~ 
    \begin{subfigure}[h]{0.3\textwidth}
        \begin{center}
        \includegraphics[width=\columnwidth]{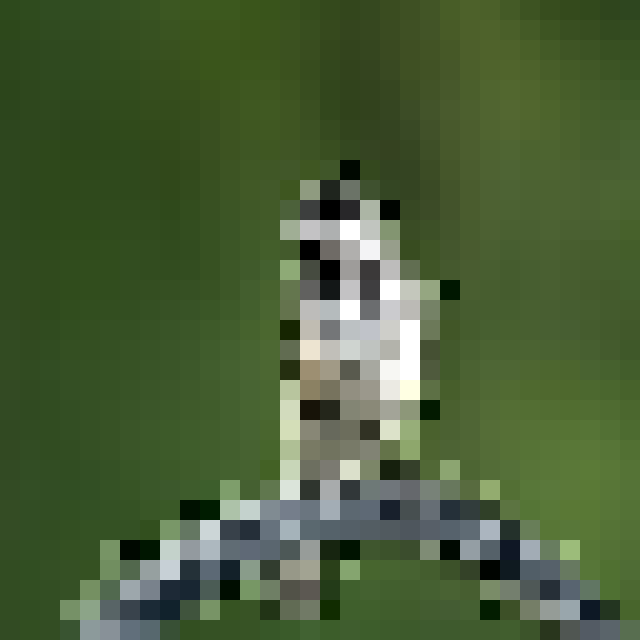}
        \end{center}
    \end{subfigure}
    \begin{subfigure}[h]{0.3\textwidth}
        \begin{center}
        \includegraphics[width=\columnwidth]{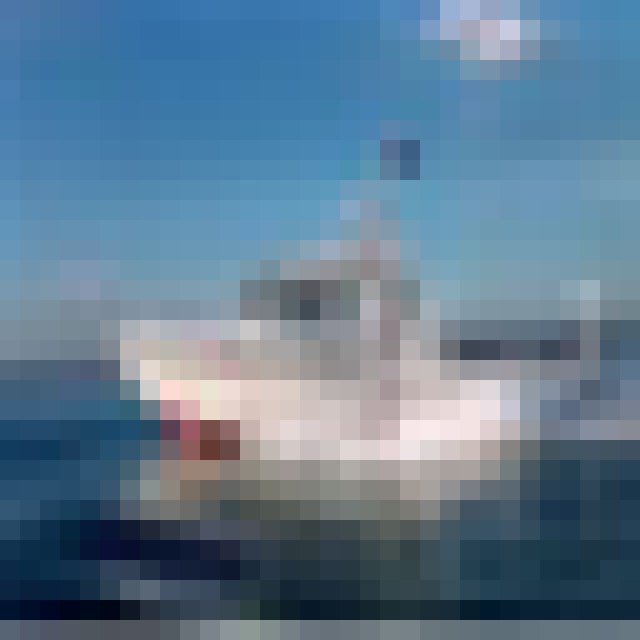}
        \end{center}
        \caption{Original images.}
    \end{subfigure}
    ~ 
    \begin{subfigure}[h]{0.3\textwidth}
        \begin{center}
        \includegraphics[width=\columnwidth]{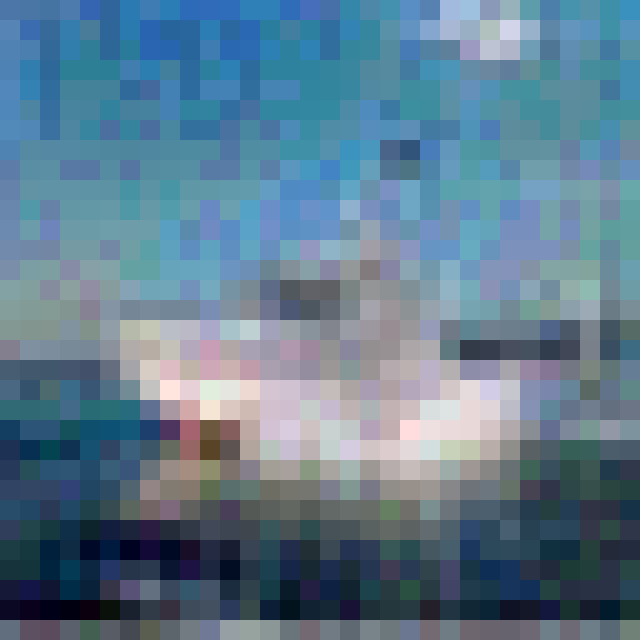}
        \end{center}
        \caption{FGSM attack.}
    \end{subfigure}
    ~ 
    \begin{subfigure}[h]{0.3\textwidth}
        \begin{center}
        \includegraphics[width=\columnwidth]{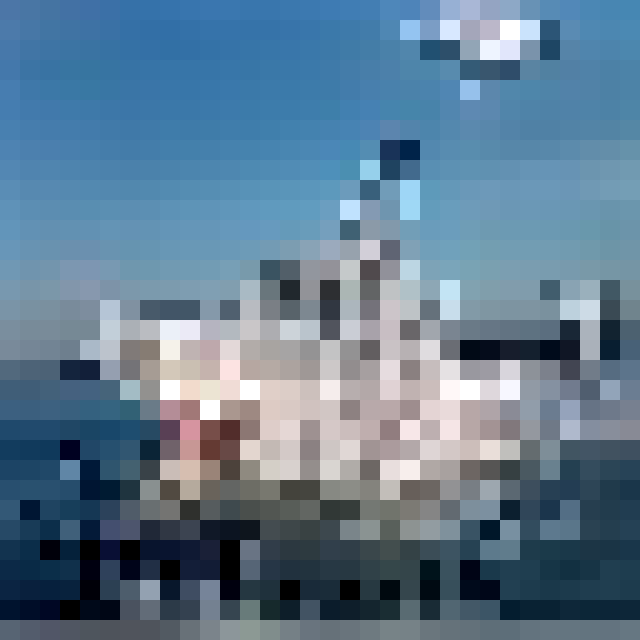}
        \end{center}
        \caption{HVS2 attack.}
    \end{subfigure}
    \caption{The bad results. The clean images in column (a) don't have enough high frequency areas to hide adversarial pixels. The HVS2 attacked images in column (c) attempt to place adversarial pixels in the few low frequency pixels, leading to large perturbations. Original image size 32x32. }
    \label{fig:bad}
\end{figure}

We found that the majority of the FGSM adversarial examples were indistinguishable. For the handful of FGSM examples that were distinguishable, the HVS2 attack would sometimes successfully generate better images (less perceptual distance) images. Other times, it would generate worse images (more perceptual distance). See Figures \ref{fig:good} and \ref{fig:bad} for examples of each. 

\section{Other approaches that failed}

Our initial hypothesis on luma and chroma was to convert the image pixels from RGB to YUV, an HSV oriented colormap that separate luma (Y) from chroma (U and V). To implement that hypothesis, we converted the gradients generated by FGSM into YUV space, then clipped to zero any perturbations to the Y (luma) channel. We used the Tensorflow implementation \citep{tensorflow2015-whitepaper}, which uses the matrices in Equations \ref{eq:yuv} and \ref{eq:rgb}. To apply the YUV perturbations, we would take our RGB image, generate a YUV image, apply the YUV perturbations, then convert back to RGB.

\begin{equation} \label{eq:yuv}
\begin{bmatrix}
    Y \\
    U \\
    V
\end{bmatrix}
= 
\begin{bmatrix}
    0.299 & 0.587 & 0.114 \\
    -0.14714119 & -0.28886916 & 0.43601035 \\
    0.61497538 & -0.51496512 &  -0.10001026
\end{bmatrix} 
\begin{bmatrix}
    R \\
    G \\
    B
\end{bmatrix} 
\end{equation}

\begin{equation}\label{eq:rgb}
\begin{bmatrix}
    R \\
    G \\
    B
\end{bmatrix}
= 
\begin{bmatrix}
    1 & 0 & 1.13988303 \\
    1 & -0.394642334 & -0.58062185 \\
    1 & 2.03206185  & 0
\end{bmatrix} 
\begin{bmatrix}
    Y \\
    U \\
    V
\end{bmatrix} 
\end{equation}

Using this approach, FGSM was generally not able to find an adversarial example. We hypothesize that the conversion between RGB and YUV acts as a hash function, reducing the overall effect of any perturbation on a DCNN trained on RGB images. 

\begin{figure}[t]
    \centering
    \begin{subfigure}[t]{0.25\textwidth}
        \begin{center}
        \includegraphics[width=\columnwidth]{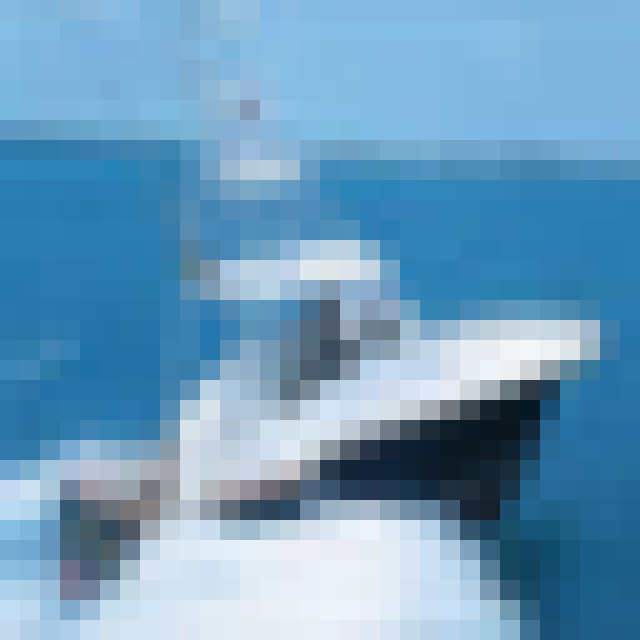}
        \end{center}
        \caption{Original image.}
    \end{subfigure}
    ~ 
    \begin{subfigure}[t]{0.25\textwidth}
        \begin{center}
        \includegraphics[width=\columnwidth]{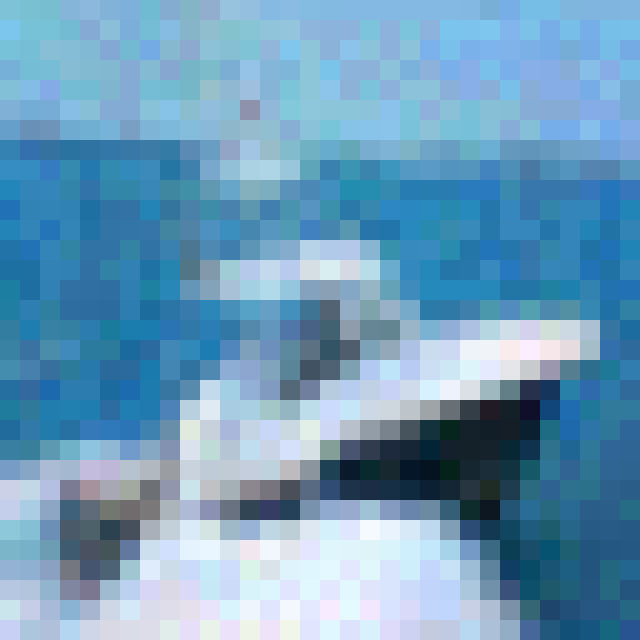}
        \end{center}
        \caption{FGSM attack.}
    \end{subfigure}
    ~ 
    \begin{subfigure}[t]{0.25\textwidth}
        \begin{center}
        \includegraphics[width=\columnwidth]{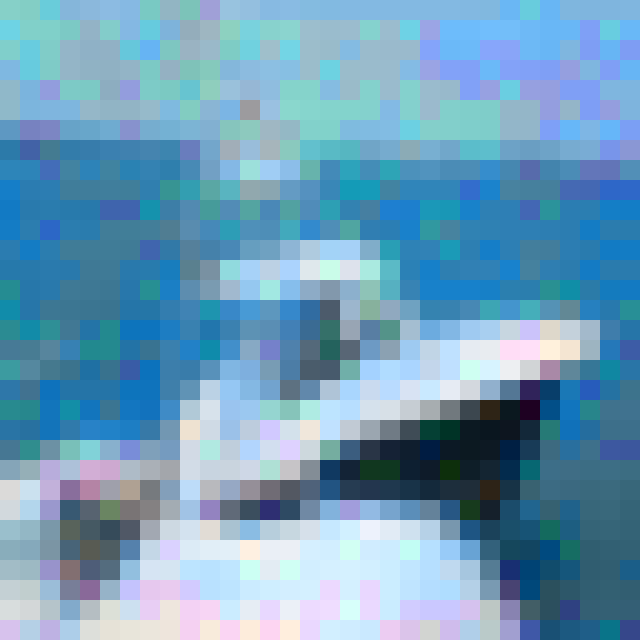}
        \end{center}
        \caption{Approximate Constant Luma attack. }
    \end{subfigure}
    \caption{Example of results from an approximate constant luma attack. The "rainbow snow" texture of Approximate Constant Luma attacked image (c) is worse than the FGSM attacked image (b). 
    Original image size 32x32. }
    \label{fig:mod1}
\end{figure}

Our second approach was to approximate our constant luma approach by searching for pixels where one of the three RGB channels was positive and one was negative. Obviously, this approach ignores clipping as well as the relatively higher luma of green pixels and lower luma of blue pixels. This attack created colorized textures that created more perceptual distance. See Figure \ref{fig:mod1}. However, we hypothesized that perhaps the theory of luma and chroma needed to be revised in the context of Adversarial AI. While changes in chroma may indeed be less noticeable to the human eye than luma, changing luma but retaining chroma would generally create perturbations within the same color palette which would be easier to hide in high frequency areas. The result was the perturbations described in the previous section. 

\section{Conclusion and future work}

In this paper, we have modified adversarial AI attacks to incorporate HVS theories of perceptual distance. We have found that simple approaches can be effective at yielding images that are less detectable by the human visual system. By design, these approaches are simple and unoptimized. We believe better results are possible through many directions for future work.

Many existing models of the HVS can be used in lieu of $L_p$ norms to optimize for a weighted average of both misclassification and perceptual distance using PGD \citep{nadenau2000human}. Even the common approach of DCT will likely outperform our simple measure of frequency \citep{JPEG}. 

Continuous clipping functions can be used for for adversarial perturbations. Our current approach essentially clips away all adversarial perturbation outside of known regions. Instead, we could allow smaller perturbations in lower frequency areas and larger perturbations in higher frequency areas.

Because people are accustomed to JPEG compression artifacts, it may be possible to hide perturbations even in low frequency areas of images if boxed in by 8x8 pixel regions to simulate the artifacts of JPEG compression. 

Existing HVS models were mainly focused on quality of image compression. There may be different HVS models for hiding adversarial perturbations. To promote further research, new mathematically HVS models focused on hiding adversarial perturbations can be developed. Initially these models will need to be benchmarked by human subjects, just like existing HVS models  \citep{sheikh2006statistical}. 

DCNNs trained on on HVS-based colorspaces like YUV may be require larger perceptual distance to generate adversarial perturbations.

Finally, outside the field of Adversarial AI, the relatively importance of luma suggests a new design for convolutional neural networks, where the luma channel has relatively more hidden layers and more weights than the chroma channels. In an extreme model, chroma could be eliminated entirely to train a black and white only DCNN.  


\acks{We would like to acknowledge support for this project from Tom Rikert, Chiara Cerini,
David Wu, Bjorn Eriksson, and Israel Niezen. }


\newpage

\vskip 0.2in
\bibliography{sample}
\end{document}